\documentclass[runningheads]{llncs}
\usepackage{times}
\usepackage{epsfig}
\usepackage{graphicx}
\usepackage{amssymb}
\usepackage{tabularx}
\usepackage{booktabs}
\usepackage{rotating}
\usepackage{array}
\usepackage{subfig}
\usepackage{placeins}
\usepackage[numbers]{natbib}
\usepackage{eucal}
\usepackage{subfig}
\usepackage{adjustbox}
\usepackage{booktabs}
\usepackage{tikz}
\usepackage{tcolorbox}
\usetikzlibrary{shapes.misc}
\usepackage{xspace}
\usepackage{natbib}
\usepackage{}

\newcommand{\specialcell}[2][c]{%
          \begin{tabular}[#1]{@{}l@{}}#2\end{tabular}}

\usepackage[pagebackref=true,letterpaper=true,colorlinks=true,bookmarks=false]{hyperref}
\usepackage{multirow}
\usepackage{amsmath}

\begin{document}
	\pagestyle{headings}
	\mainmatter

	\def\GCPR16SubNumber{90}

	\title{From Traditional to Modern : Domain Adaptation for Action Classification in Short Social Video Clips}

	\titlerunning{Domain Adaptation for Action Classification in Short Social Video Clips}
	\author{Aditya Singh \and Saurabh Saini \and Rajvi Shah \and P J Narayanan}
	\institute{Center for Visual Information Technology, IIIT Hyderabad, India}

	\maketitle

\begin{abstract}
Short internet video clips like \textit{vines} present a significantly wild distribution compared to traditional 
video datasets. In this paper, we focus on the problem of unsupervised action classification in wild vines using 
traditional labeled datasets. To this end, we use a data augmentation based simple domain adaptation strategy. 
We utilize semantic \textit{word2vec} space as a common subspace to embed video features from both, labeled source domain 
and unlabled target domain. Our method incrementally augments the labeled source with target samples and 
iteratively modifies the embedding function to bring the source and target distributions together. 
Additionally, we utilize a multi-modal representation that incorporates noisy semantic information 
available in form of hash-tags. We show the effectiveness of this simple adaptation technique on a 
test set of vines and achieve notable improvements in performance.

\end{abstract}

%
\section{Introduction}
\label{sec:intro}
Action classification is an active field of research due to its applications 
in multiple domains. The last decade has seen a significant paradigm shift from 
model-based to data-driven learning for this task. Over the years, increasingly complex and challenging action 
recognition datasets such as UCF101, HMDB, Hollywood, etc.\ have been introduced 
\cite{ActionsAsSpaceTimeShapes_pami07, YouTubeAction, OlySports, HMDB, hollywood, UCF101, 2015trecvidover}. 
However, with growing popularity of 
social media platforms and mobile camera devices, there is unprecedented amount 
of amateur footage that is significantly wilder and complex than curated datasets. 
In this paper, we analyze this problem on a particular distribution of short video 
clips shared on the social media platform \url{vine.co}, known as \textit{vines}. 
Vines are six second long, often captured by hand-held or wearable devices, with 
cuts and edits, and present a significantly wilder and more challenging 
distribution. 
 The action classifiers trained using traditional distributions 
such as UCF, HMDB, etc.\ cannot generalize or adapt to wild 
distributions like vines \cite{EfrosBias, KhoslaBias, SultaniBias}. 
Recent methods that use increasingly complex features from large-scale dictionaries and 
Convolutional Neural Networks (CNN) are showing promise in building more generalizable systems. 
However, these methods require supervised training with large-scale labeled data. 
Moreover, data from Internet sources, like  vines, is ever 
increasing and manually labeling such data is tedious and expensive. 
A better approach is to expand the scope of existing datasets and classifiers.

\begin{figure}[t]
\vspace{-2mm}
\centering
\subfloat[]{\includegraphics[width=0.45\textwidth,height=5cm]{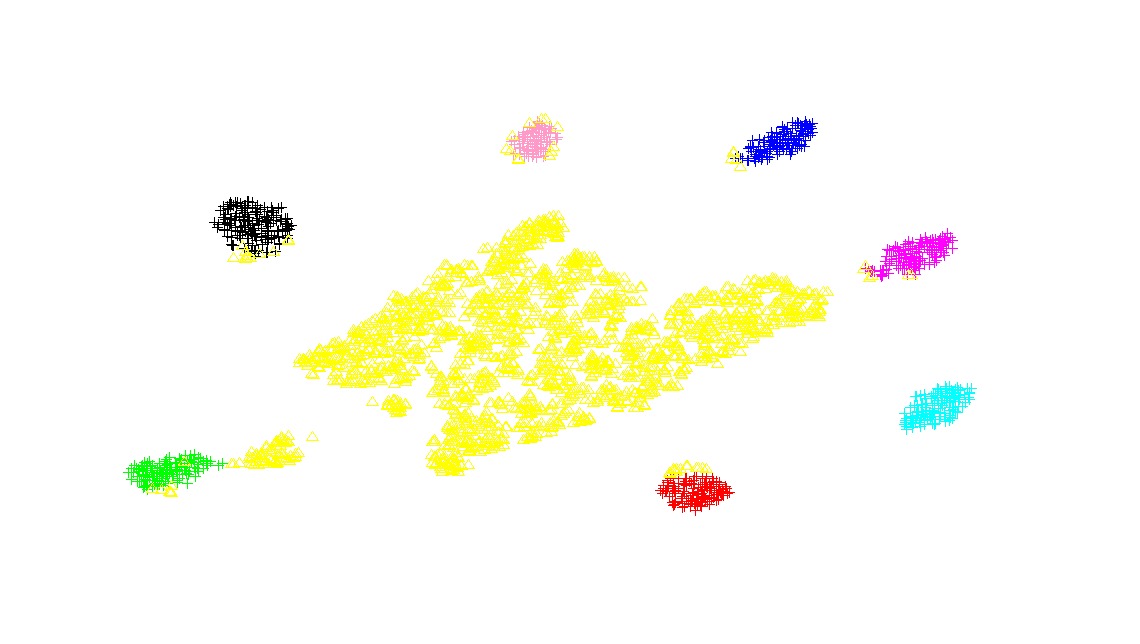}}
\subfloat[]{\includegraphics[width=0.45\textwidth,height=5cm]{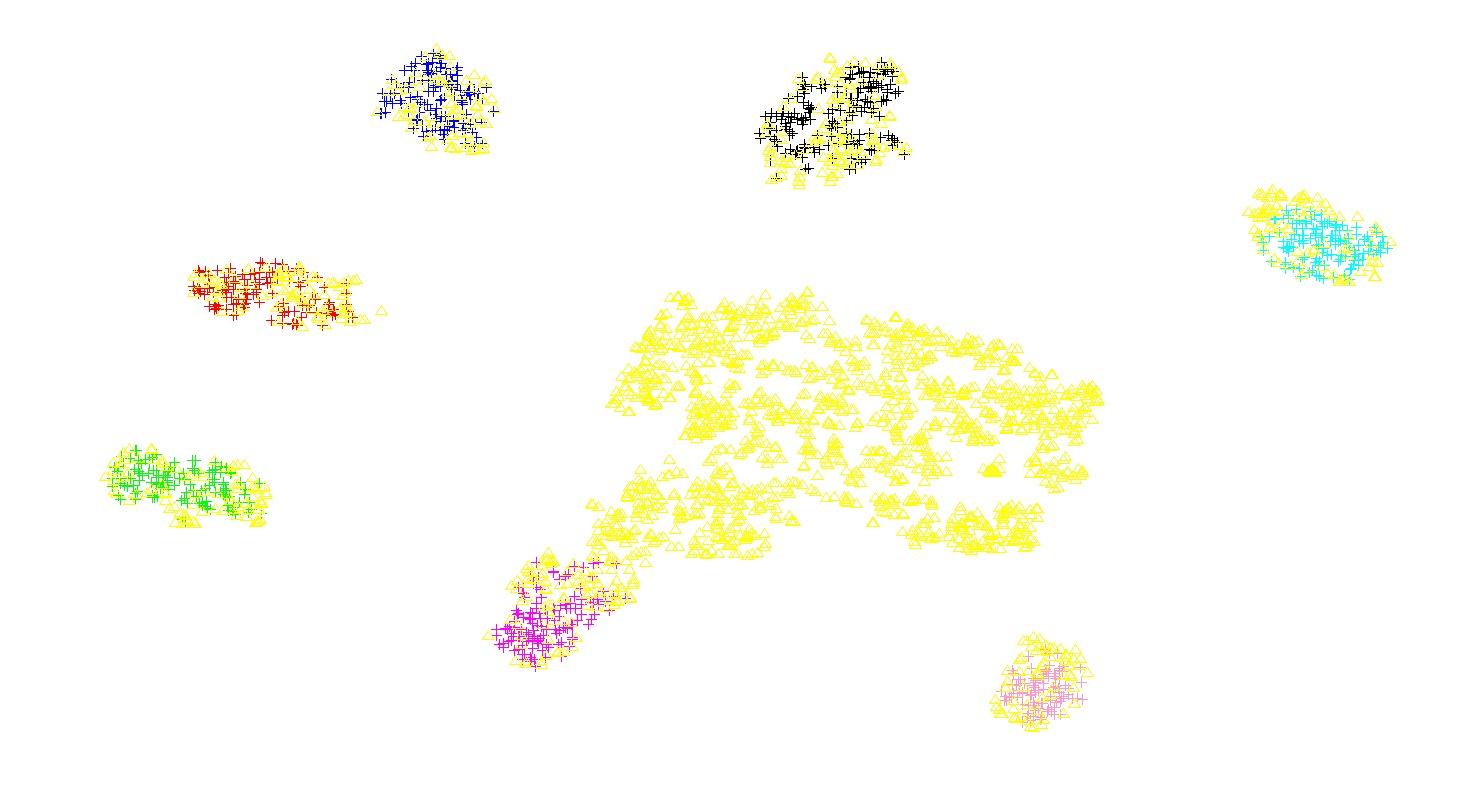}}
\caption{t-SNE Visualization of semantic embedding of UCF and Vines before and after iterative training. 
The crosses represent auxiliary (UCF) samples and are color-coded according to their class labels. The 
yellow triangles represent unlabeled target training samples (vines). 
After iterations, many 
more vines merge with the clusters formed by the auxiliary samples. The leftover vines possibly 
belong to none of the action classes and hence do not merge into any cluster. Please see this visualization in color.}
\vspace{-4mm}
 \label{fig:vis_two_distributions}
\end{figure}

We present a simple incremental approach to transfer knowledge from a traditional 
labeled source dataset to a wilder unlabeled target dataset. The class of methods that try to mitigate the bias/shift between 
different distributions/datasets fall under the category of \textit{transfer learning} methods. 
Recent approaches along these lines include works on dataset bias shift \cite{EfrosBias,datasetIssues}, domain adaptation \cite{adaptiveDict,whatYouSeeDomain, acRecDA1}, 
zero-shot learning \cite{zslearn,ZSLRegression}, heterogeneous/multi-modal transfer \cite{NIPS2008_3550} and other transfer learning methods 
\cite{daumeiii,GFK,YangDA,5288526}. However, most of these methods work on image/object category problems or 
text-data problems and their application to web-scale wild video distributions remain untested. 
\citet{YouTubeCat} use hierarchical category taxonomy tree, designed by professional linguists, to categorize 
Youtube videos. However, this approach cannot be extended for action classification as it is difficult to hand-craft a generalized 
taxonomy for actions. \citet{SultaniBias} propose a feature encoding that accounts for the bias 
introduced by dataset specific backgrounds for video classification. However, this method requires both 
source and target videos to be labeled.

We propose a simple, unsupervised approach that iteratively adapts the base classifiers 
trained on a labeled training set UCF50 to an unseen, unlabeled set of vines, by 
incrementally augmenting the training set with vines. Adopting the terminology of 
domain adaptation and transfer learning literature, we call the UCF50 labeled set 
an \textit{auxiliary training set} and the unlabeled vines a \textit{target training set}. 
We leverage a semantic space \textit{word2vec} \cite{w2v} as a common reference space 
to bring together the auxiliary and the target domains. To embed video features in 
this space, we learn a neural-network based embedding function \cite{zslearn}. 
We first learn this embedding using labeled samples of the auxiliary set and project both labeled and 
unlabeled samples from auxiliary and target sets into the semantic space. \autoref{fig:vis_two_distributions}(a) shows a 
t-Distributed Stochastic Neighbor Embedding (t-SNE) visualization of seven classes after projection 
into the semantic space. The auxiliary samples are represented by crosses, color-coded as per their class 
labels; the target samples are represented by yellow triangles. It can be clearly seen that the auxiliary and target domains are disparate. While auxiliary training 
samples (UCF) form separable clusters, most target training samples (vines) are cluttered and inseparable in the semantic space. 
Though it is not possible to reliably classify all target samples in this space, we use a multi-modal scoring function to 
select a few vine samples from the target that can be classified with high confidence. Our multi-modal scoring function also 
incorporates the knowledge from user-given hash-tags for classification. We add these samples to the 
labeled auxiliary training set, and retrain the embedding function using the augmented auxiliary training set. After several 
iterations of this process, auxiliary set is augmented with sufficient samples from the target distribution. 
\autoref{fig:vis_two_distributions}(b) shows the t-SNE visualization of the embedding after several 
iterations of augmentation and retraining. It can be seen that, after iterations, 
many more target samples merge with the clusters formed by the auxiliary samples. 
\enlargethispage{\baselineskip}

A recent work \cite{7350760} also leverages semanatic embedding for recognizing new action categories in a zero-shot learning framework for traditional datasets (UCF and HMDB). However, the focus of our work is on  learning cross-domain action classification for wild social web-videos. Also, our method incrementally relearns the neural network allowing more non-linearity in the embedding function and utilizes 
multi-modal features that include motion features and hash-tags. The nature of this work is experimental and exploratory. We perform experiments for 7 action classes of UCF50 dataset and show that this surprisingly simple strategy works effectively and yields precision, recall, and F-measure improvement of 2\% to 10\% on an unseen vines test set. Please visit our web-page for more information and research resources,\\ \texttt{\url{https://cvit.iiit.ac.in/projects/actionvines/}}.
%
%

%
\section {Our Approach}
\label{sec:method}
\begin{figure}
 \fboxsep=2mm
\fboxrule=2pt
\resizebox{\textwidth}{!}{%
\begin{tabular}{l}
{\rotatebox{90}{\hspace{4mm}\textbf{cycling}}}
\fcolorbox{green}{white}{
\includegraphics[height=2.2cm]{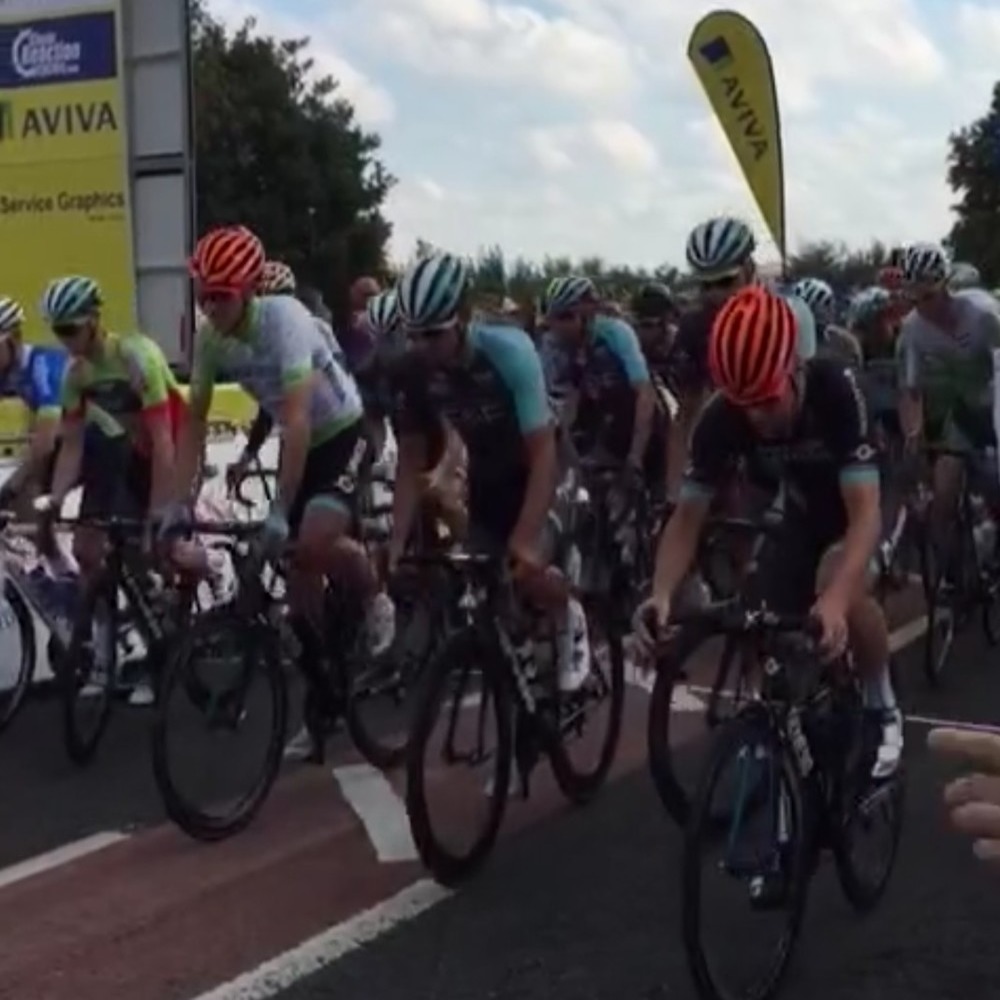}\,
\includegraphics[height=2.2cm]{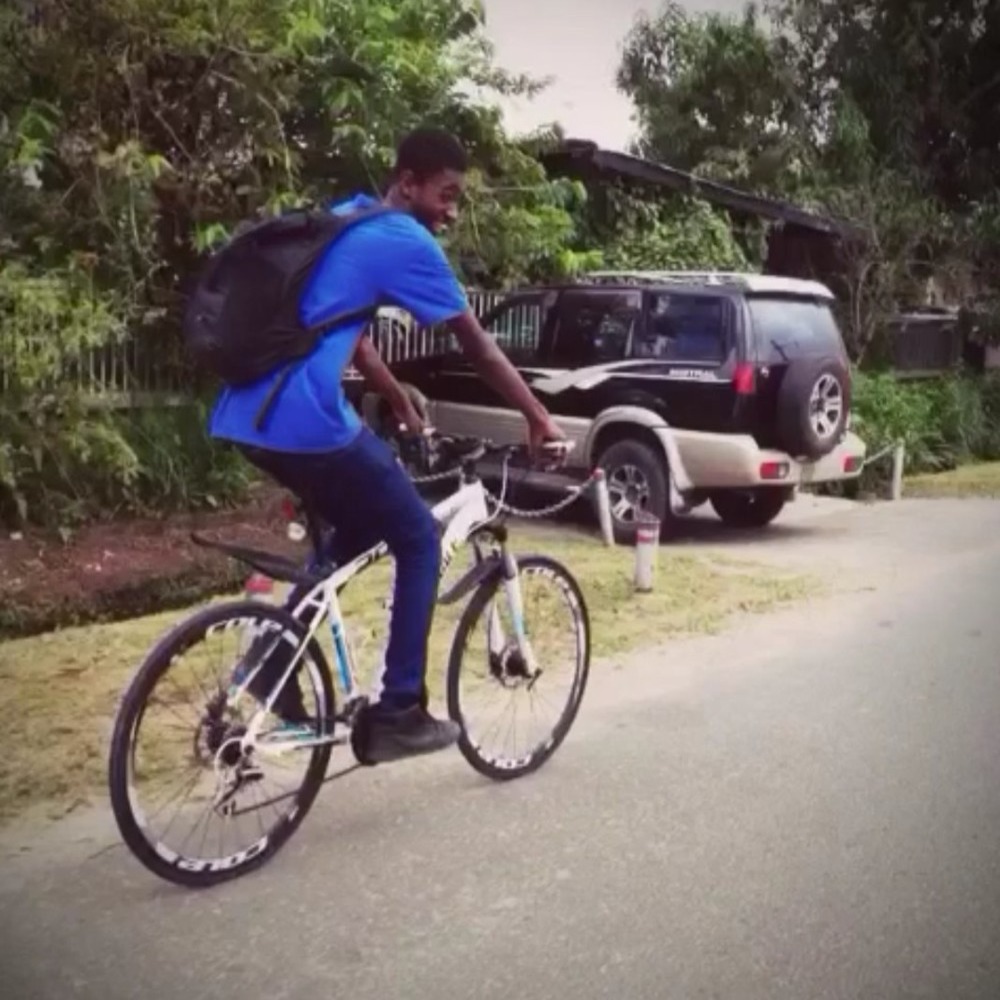}\,
\includegraphics[height=2.2cm]{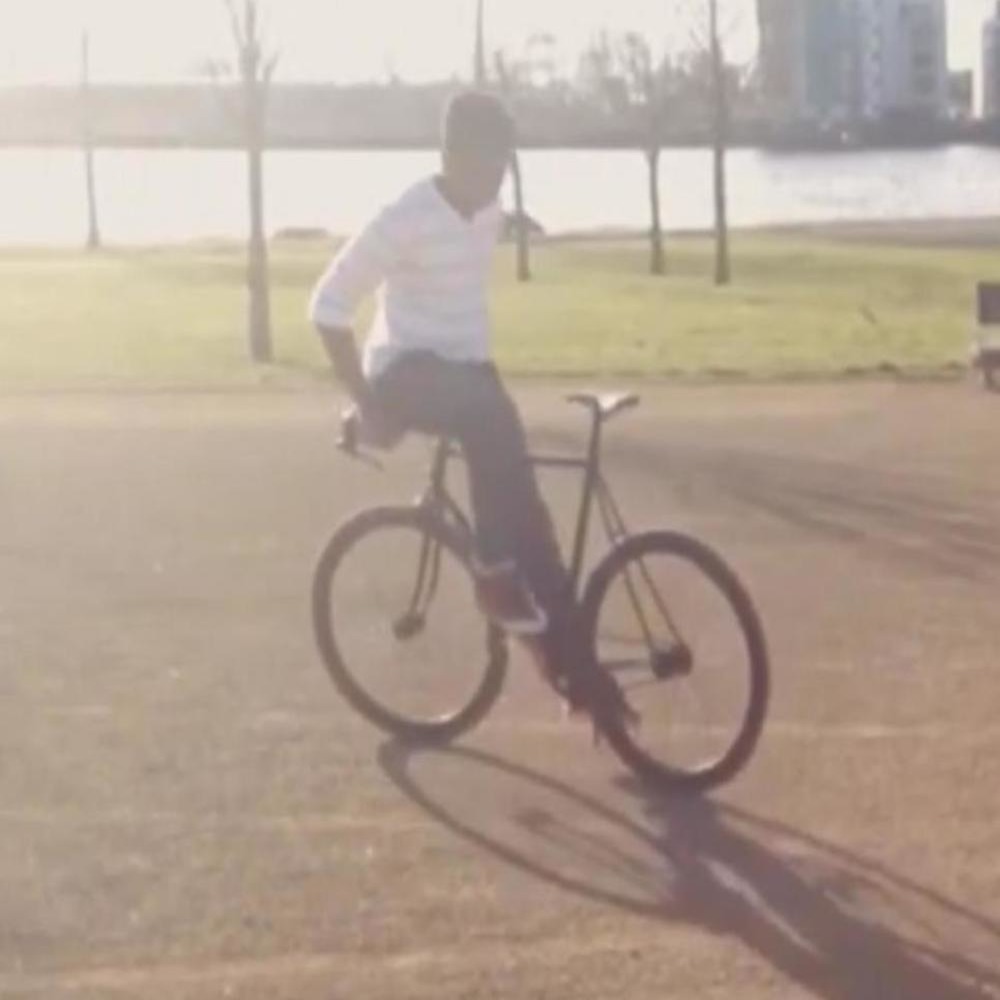}
}\hspace{0.5mm}
\fcolorbox{red}{white}{
\includegraphics[height=2.2cm]{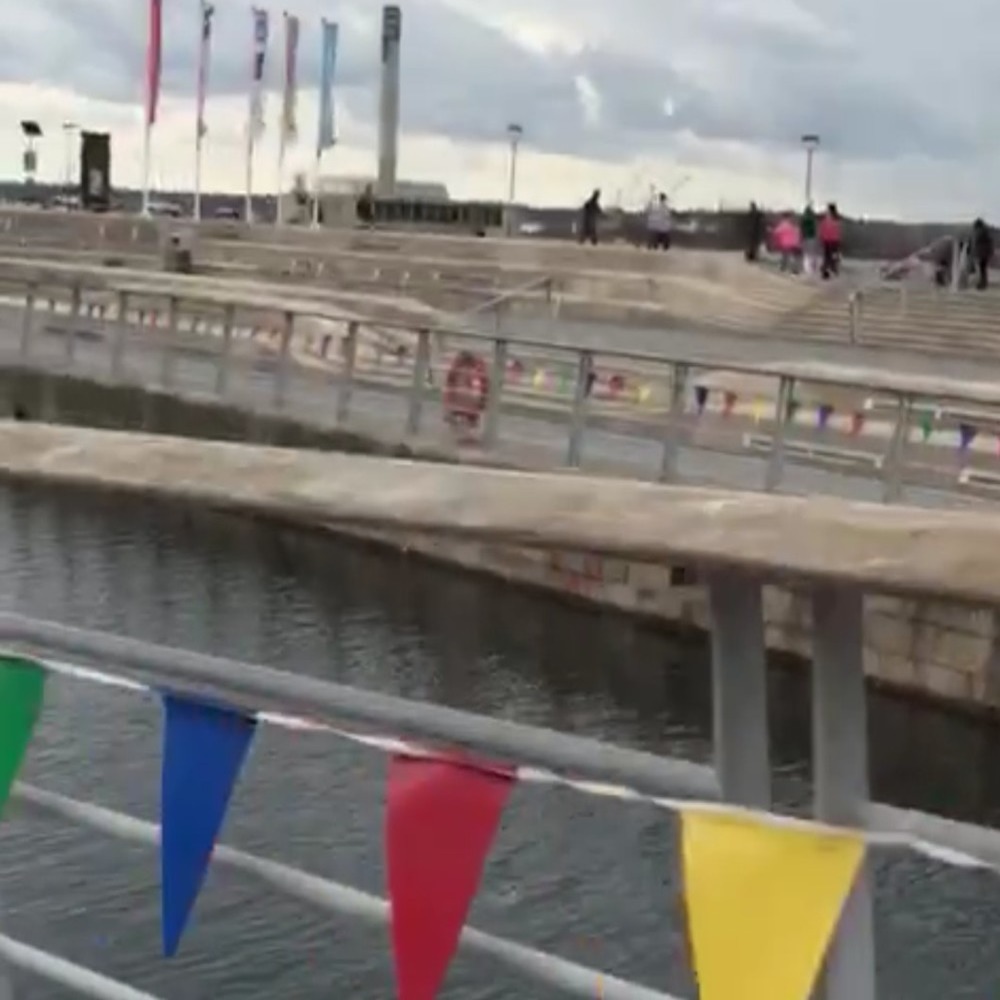}\,
\includegraphics[height=2.2cm]{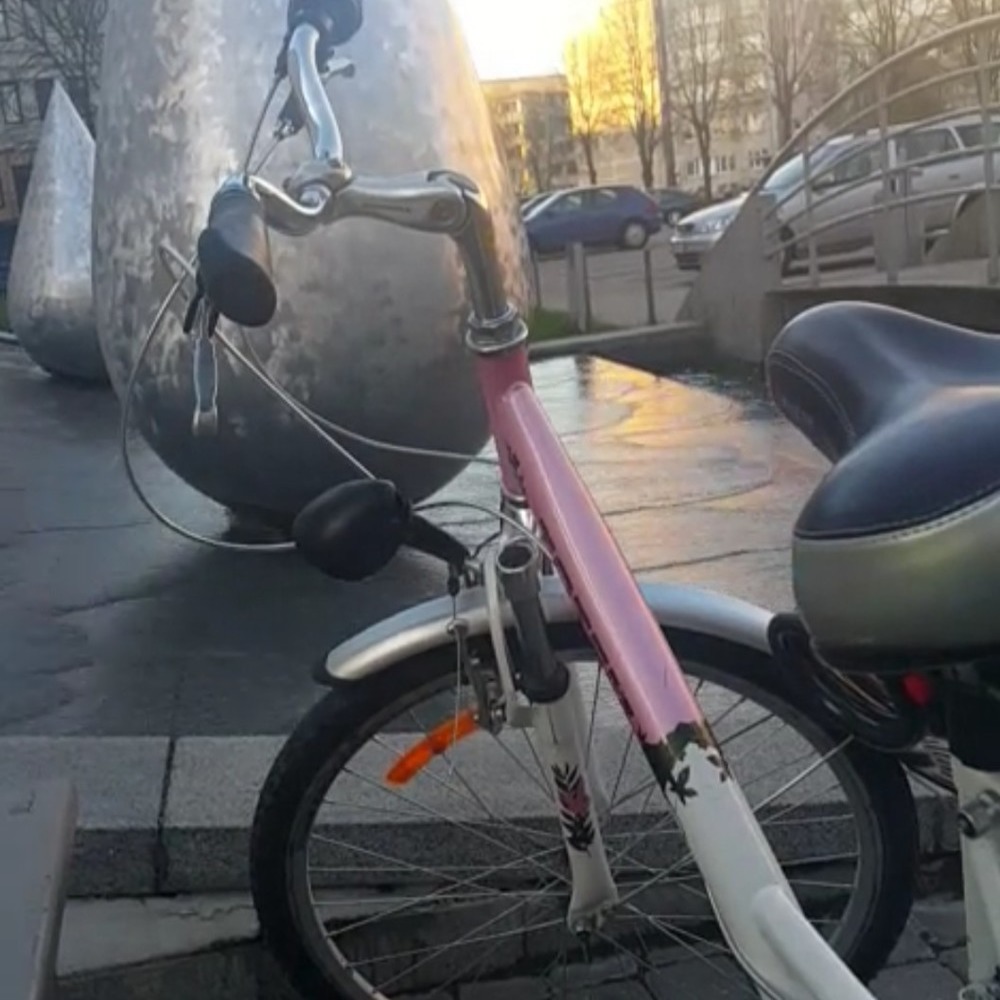}\,
\includegraphics[height=2.2cm]{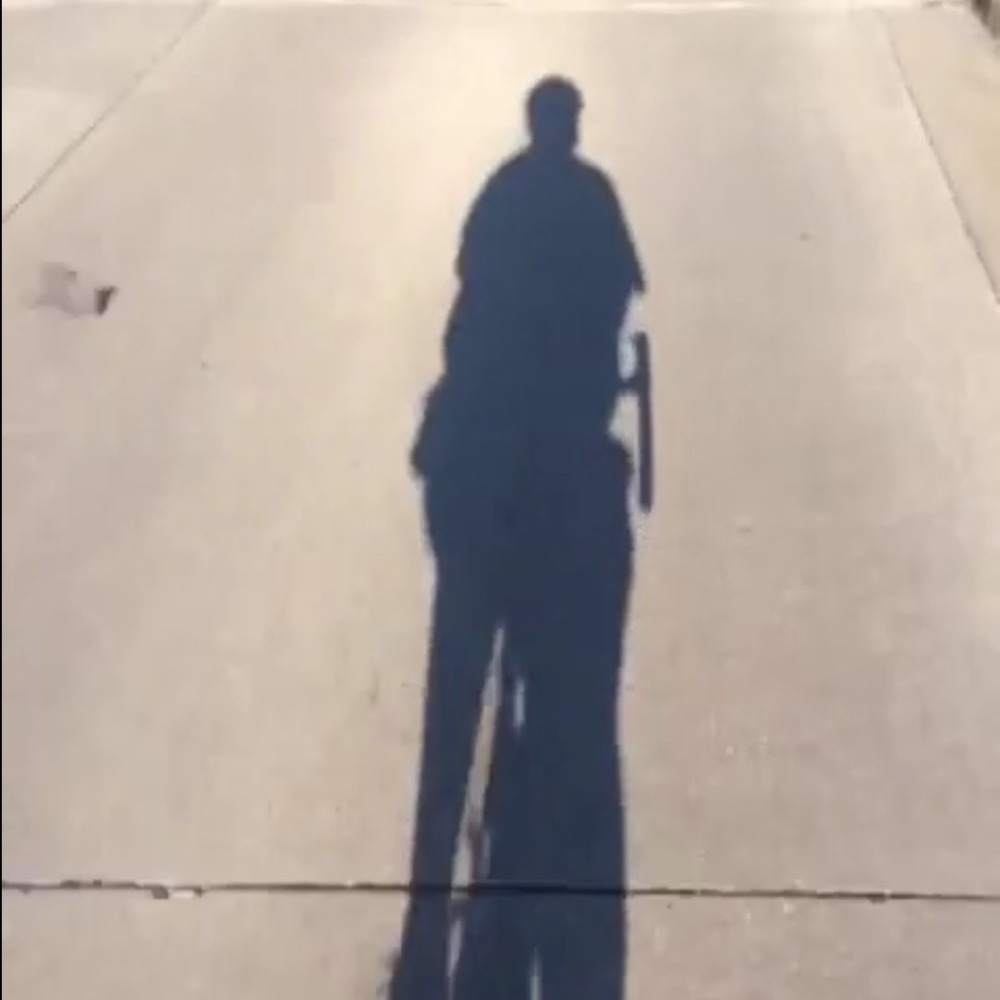}\,
\includegraphics[height=2.2cm]{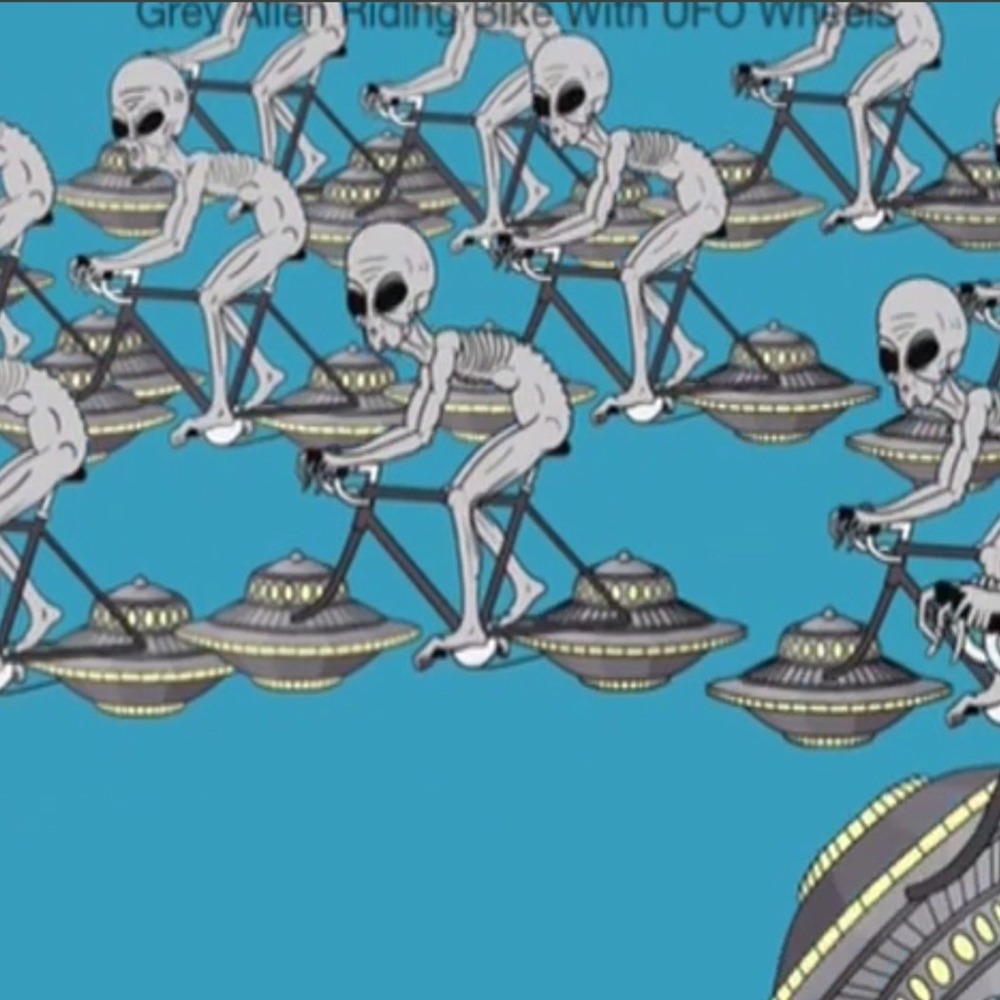}}\\[2mm]
\rotatebox{90}{\hspace{4mm}\textbf{diving}}
\fcolorbox{red}{white}{
\includegraphics[height=2.2cm]{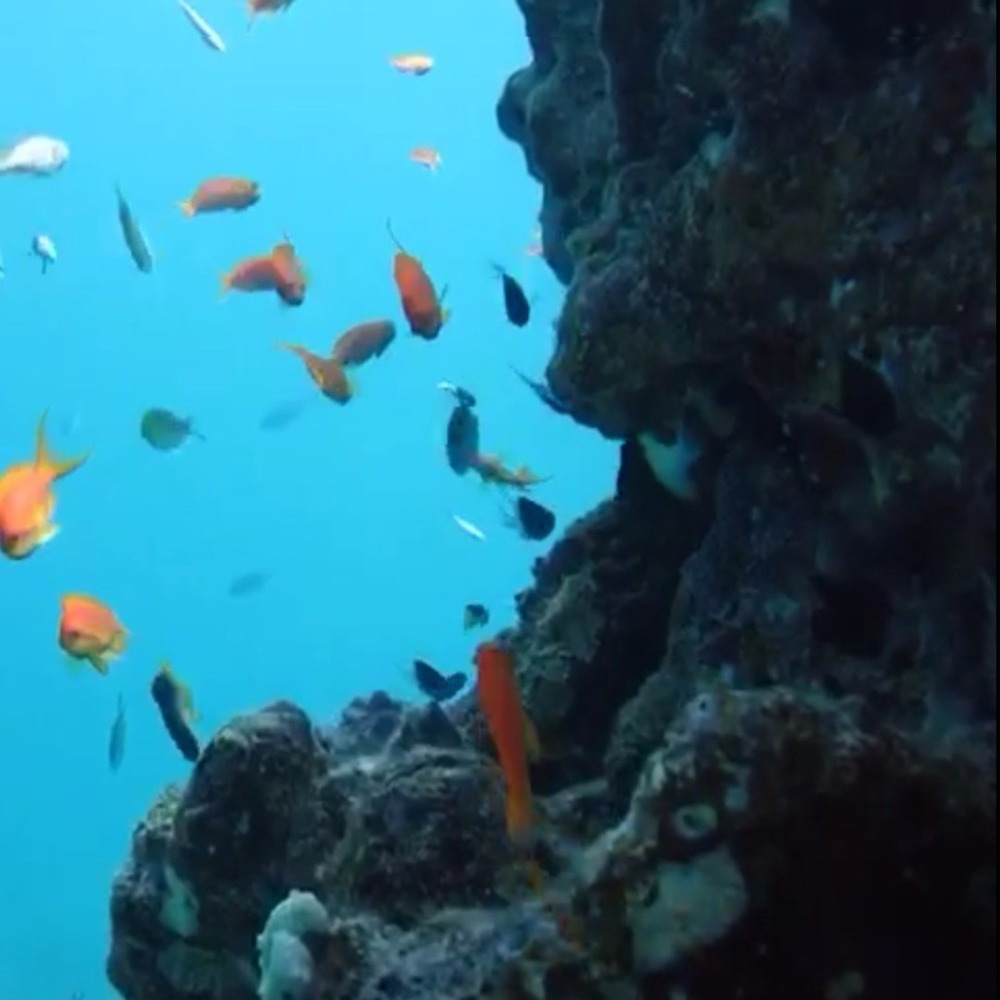}\,
\includegraphics[height=2.2cm]{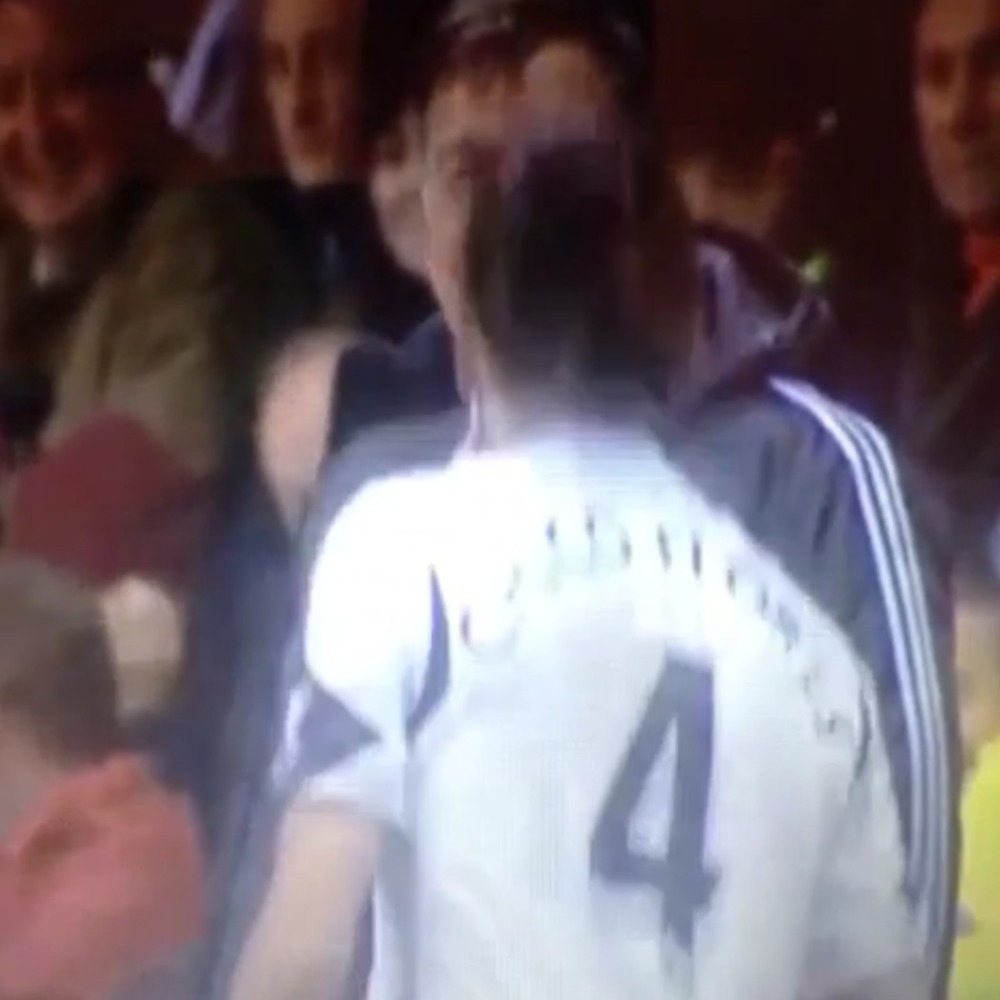}\,
\includegraphics[height=2.2cm]{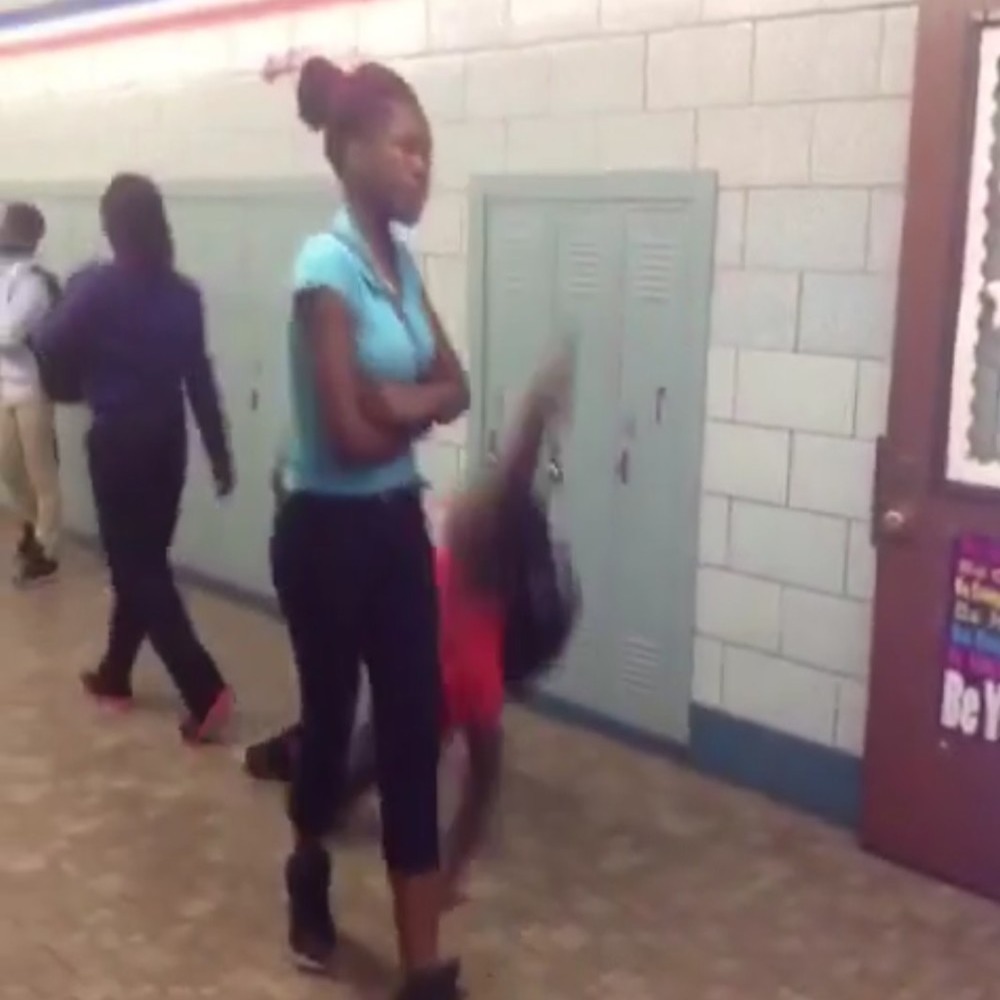}
}\hspace{0.5mm}
\fcolorbox{green}{white}{
\includegraphics[height=2.2cm]{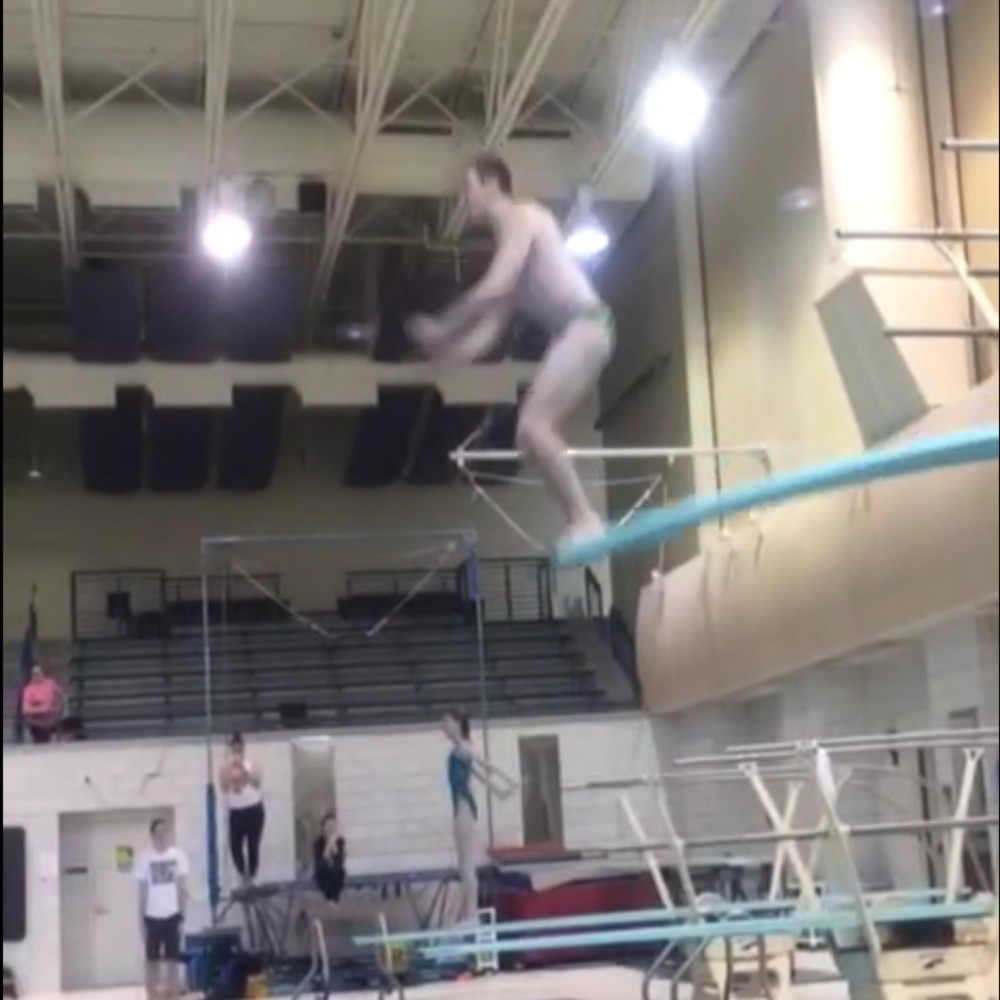}\,
\includegraphics[height=2.2cm]{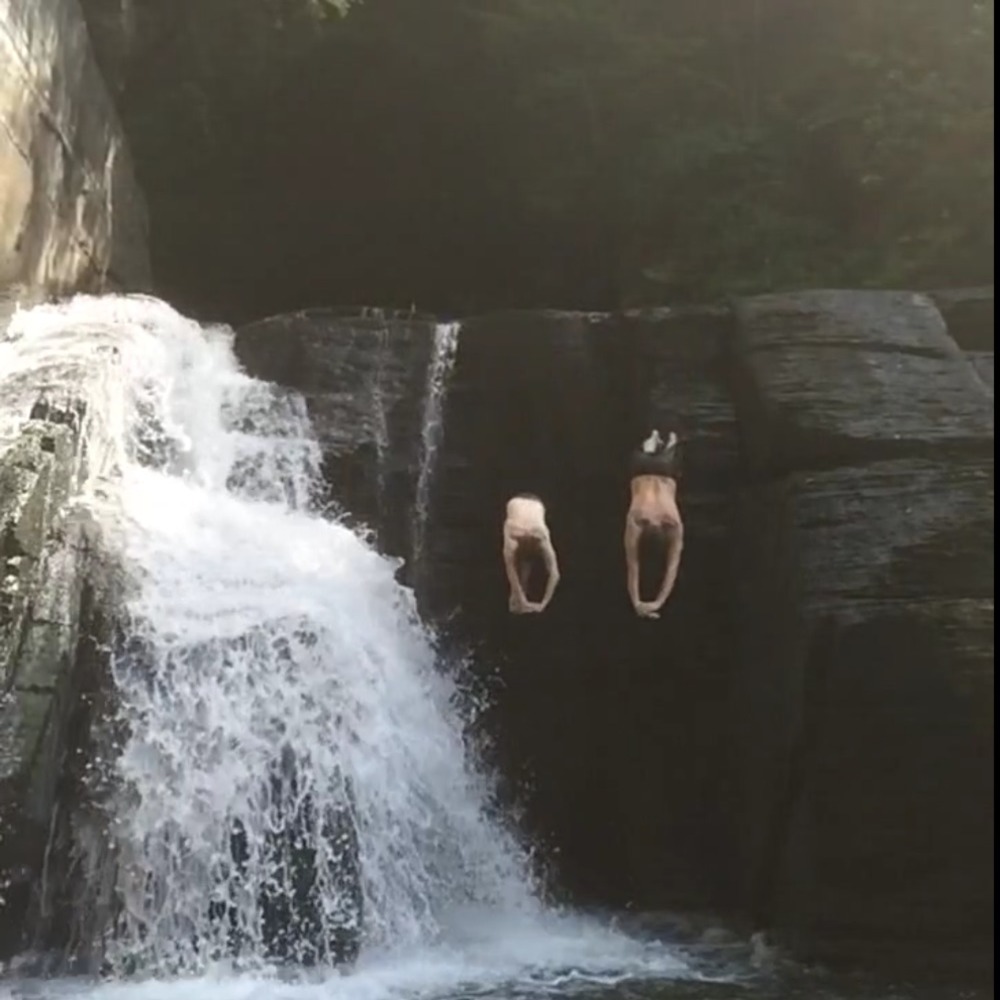}\,
\includegraphics[height=2.2cm]{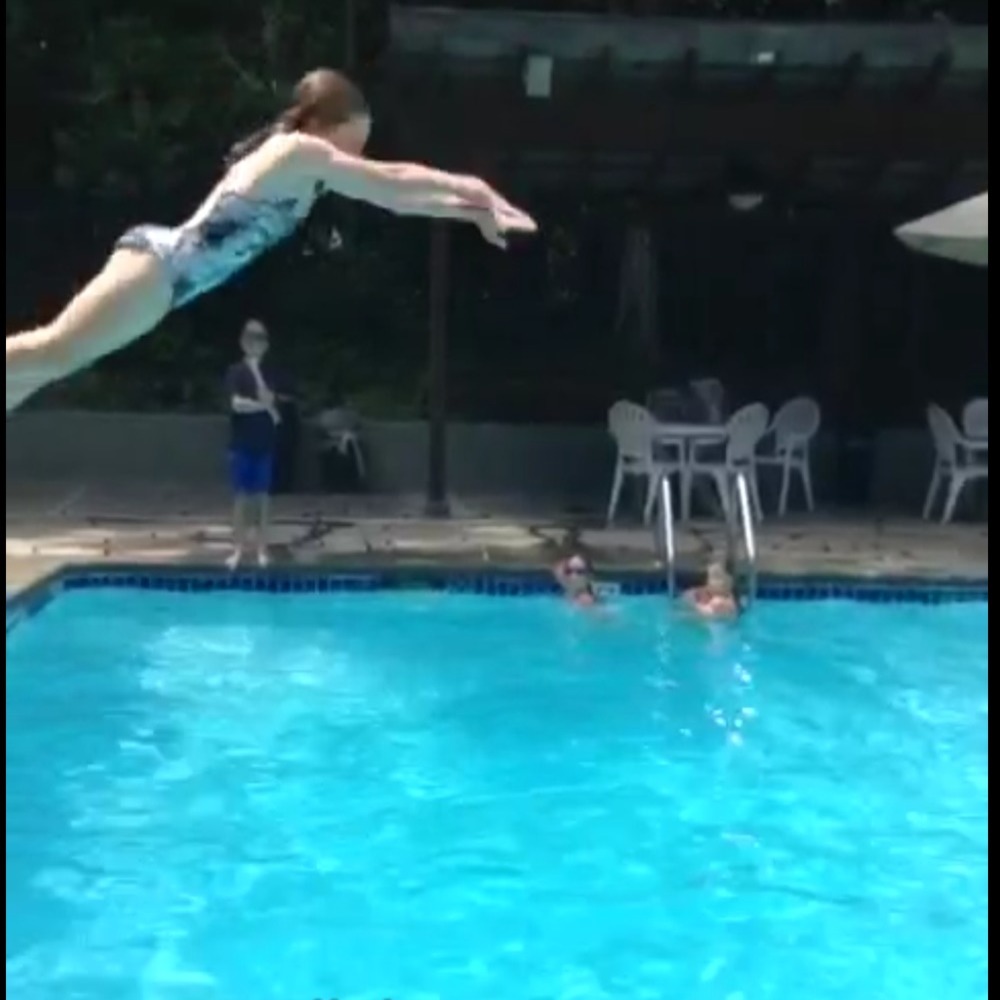}\,
\includegraphics[height=2.2cm]{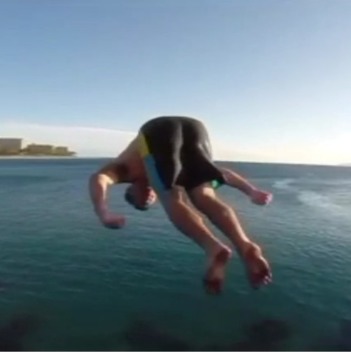}}\\
\\
\end{tabular}
}
\caption{Stills from sample vines (short video clips) retrieved for two queries, `cycling' (top) and `diving' (bottom). The vines in the red boxes are semantically related concepts to the query words but are negative examples for the query human action. The vines in the green boxes are positive examples for the respective human actions. The high intra-class variability is worth noting.}
\label{fig:sample_frames}
\end{figure}

The distribution of vines is widely different from traditional action classification datasets in terms of appearance, quality, 
content, editing, etc. For a classifier to work well for vines, it needs to be trained on a labeled set of vines. 
Except, manually annotating such ever-altering web data is tedious and impractical. Vines do come with user-given  
tags, and description but such tags cannot be considered reliable labels. 
\autoref{fig:sample_frames} shows stills from vines retrieved for two action words `cycling' 
and `diving' as queries. Though both sets of vines are hash-tagged with their respective action words, 
some vines are only related to the concept and not the human action, while some vines are completely 
unrelated to either the concept or the action. We tackle the problem of improving action classification for vines 
by utilizing labeled samples from an auxiliary domain and unlabeled samples from a target domain with noisy and 
weak semantic information. In the following subsections, we provide details of data collection, multi-modal 
feature representation, and iterative training. 

\begin{table}[h]
\setlength{\tabcolsep}{4pt}
  \resizebox{\linewidth}{!}{%
 \centering
 \begin{tabular}{l l c c c c c c c}
 \toprule
 Action Class & & Billiards & Cycling & Diving & Golfswing & Horseride & Kayaking & Push up\\
 \midrule
 \multirow{3}{*}{\specialcell{Target Domain \\(vines)}} 
 & Train (total) &267 &280 & 258& 268&233 &284 &286 \\
 & Train (true +ves) &100 &92 &133 &151 &106 &73 &178 \\
 & Test & 24&27 &39 &34 &29 &16 &40 \\
 \midrule
 \specialcell{Auxiliary Domain \\ (UCF)} & &129 &119 &124 &120 & 169&129 &90 \\ 
 \bottomrule
 \end{tabular}
 }
 \vspace{3mm}
 \caption{Number of samples in auxiliary and target sets across classes. For some 
 classes the true positives are less than $25\%$ of the total samples in the target train set. This 
 imbalance indicates the fact that `hash-tags' can only provide noisy labels and other modalities need to be 
 utilized for effectively labeling target training vines.}
\label{tab:base_datasets}
\vspace{-8mm}
\end{table}

\subsection{Data collection and statistics}
We work on seven of the fifty action categories of UCF50 action recognition dataset. 
These seven classes are selected based on the sufficient availibility of related videos on 
vine.co. The labeled samples of UCF50 belonging to these seven classes form our auxiliary 
domain/ auxiliary set and the vines form our target domain/ target set. To collect relevant 
vines for each action category, we use the action term as the query word
and download the top $450$ retrieved vines per category (restricted by the vine API). 
The retrieval of vines is based on the occurrence of the query word in either the corresponding `hash-tags' or
the description. We discard the vines that do not have the action category word
as one of the hash-tags. Thus, we have a total of 2357 vines with the associated tags. This forms 
our target domain. We divide this set into a target training set and a test set. Our incremental and iterative training for 
augmenting the auxiliary domain operates on the target training set and we report the 
performance of the final classifier on the test set. All the retrieved vines are manually annotated 
by three human operators but we never use the labels of the target training set in anyway for our training 
but only to gather data statistics, making our approach completely unsupervised.

\autoref{tab:base_datasets} shows the distribution of samples across classes and 
auxiliary, target train, and test sets. Since, the hash-tags are noisy, many vines in the target 
set do not have the respective action (false positives). The test set is pruned to remove all 
such false positives. However, since our training is unsupervised, we do not alter the target 
train set. The first two rows in \autoref{tab:base_datasets} shows the total samples in the 
target train set and the number of true positives for each class. We provide this statistic 
to demonstrate the fact that hash-tags are extremely noisy labels. This fact can also be 
observed from leftover vines in \autoref{fig:vis_two_distributions}.

%
%
%
%


\subsection{Feature Representation}
Many feature representations based on 
spatio-temporal constructs \cite{ActionsAsSpaceTimeShapes_pami07, spatioTemporalwords, Weinland:2006:FVA:1225844.1225855}, 
appropriate human body modeling \cite{Ikizler20091515, Jhuang:ICCV:2013}, successful 
image features \cite{Scovanner:2007:SDA:1291233.1291311, Klaser_aspatio-temporal, actionlets}, etc.\ are proposed in 
action recognition literature. 
Our approach leverages multi-modal feature representation to reliably augment the 
auxiliary set with target samples. We use motion features, 
semantic embedding features, and tag-distribution features in our method. 
We explain these features and the related terminology in this section. 
More details on parameters and code are given in \autoref{sec:expsetup}.

%

\paragraph{Motion features:} 
Motion encoding is the most preferred feature 
representation for action recognition training. 
We compute the fisher vector encoded improved dense trajectories (IDT) 
\cite{ImpDenseTraj} for samples of both auxiliary and target sets. 
IDT features include histogram of oriented gradients (HoG), 
histogram of optical flow (HoF), and motion boundary histogram (MBH) 
descriptors across frames. To classify these features, we use 
linear support vector machines (SVM). 

\paragraph{Semantic features:}
\citet{w2v} provide a mechanism to represent a word as a vector in a 300-dimensional vector space, 
commonly known as \emph{word2vec} space. \citet{zslearn} proposed a neural network based 
supervised method to learn a non-linear function that embeds visual (image) features into the 
word2vec space based on the corresponding object category words. We use this framework and 
learn a semantic embedding function that projects motion features (fisher vectors) into the 
word2vec space corresponding to the action word. We call the resulting 300-dimensional 
representation, embedded semantic features, or simply semantic features.

\paragraph{Tag Features:}
Hash-tags can be seen as noisy semantic labels of a video provided by the users. We assume that similar videos will have similar tags. Tagged words are usually slangs which
are used to describe a video in an informal manner and hence do not strictly adhere to the word2vec representation of word 
space. Hence, we utilize tag features in a separate framework. First, we collect all tags associated with the vines in our target dataset and perform stemming to obtain cleaner, non-redundant set of tag-words. We then create a histogram of all tag-words and form a tag dictionary by removing all singleton words. A tag feature for a vine is simply a binary vector of the dimension of the dictionary (1048 in our experiment), such that the value in $i^{th}$ position indicates whether the $i^{th}$ tag in the dictionary is associated with
the vine or not. 
\subsection{Iterative Training}

\begin{figure}
\centering
 \includegraphics[width=\linewidth]{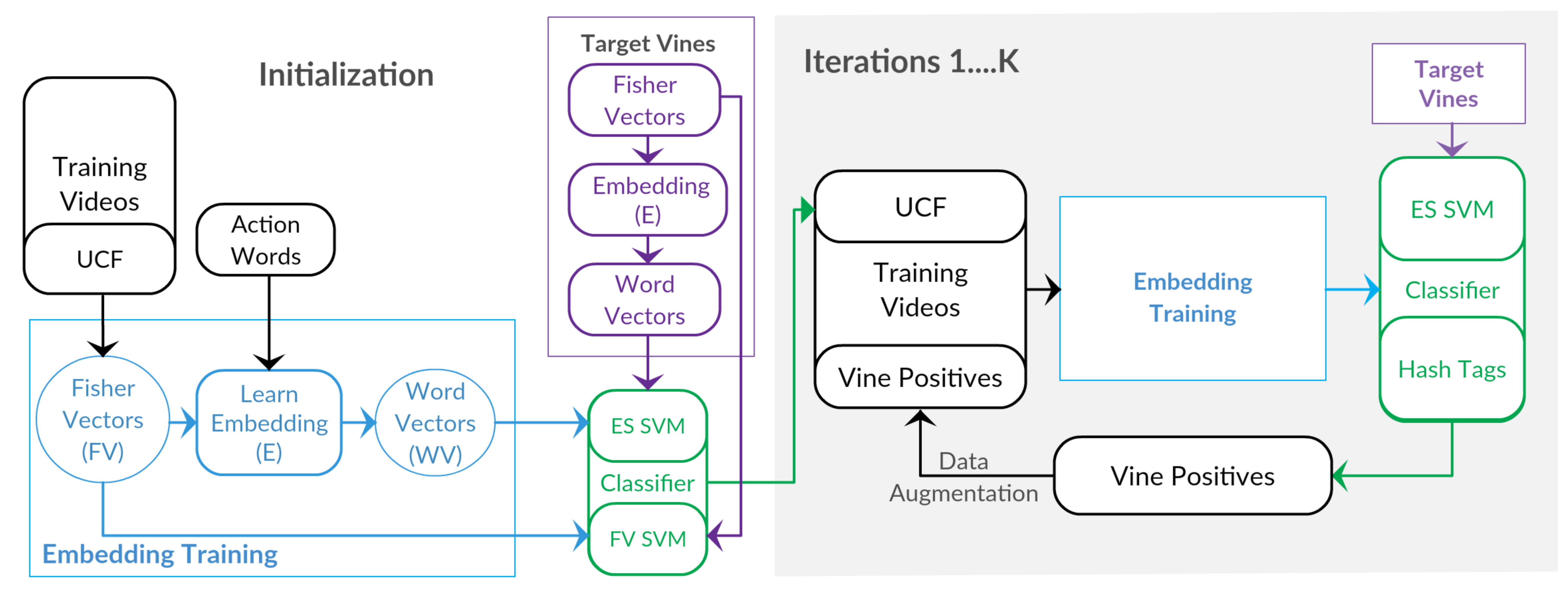}
 \caption{Block representation of our iterative training approach. Left block shows the specifics of the initialization step and the right block shows the process for iterative updates. The embedding training 
 sub-block (blue) is the same for both initialization and iteration steps (depicted in detail for only initialization step).}
 \label{fig:block_diag}
\end{figure}

\autoref{fig:block_diag} shows a block diagram of the proposed iterative training method. Each step of this method is explained in detail here. We first describe the notations used, then discuss the strategy for initialization and 
incremental updation of the training set, followed by explanation of these updation rules, and sampling choices. 

\paragraph{Notation:}
We denote the set of action categories as $\mathbb{C} = \{c_i\ \, | i \in [1,7]\}$, 
where $c_1 - c_7$ represent the seven action categories `billiards', `cycling', 
`diving', `golf', `horseriding', `kayaking', and `pushups'. 
The negative label corresponding to an action category $c_i$ is represented 
as $\tilde{c_i}$. The SVM classifiers for fisher vectors and embedded semantic features 
are denoted respectively by $\mathrm{H_{FV}}$ and $\mathrm{H_{WV}}$. 
The auxiliary set features are represented as $\mathbb{A}$.
The training set at $k^{th}$ iteration for learning the classifiers for category $c_i$ is 
denoted by $\mathbb{T}_k^{c_i}$. The sets of positive and negative examples in $\mathbb{T}_k^{c_i}$ 
are denoted respectively as $\mathbb{P}_k^{c_i}$ and $\mathbb{N}_k^{c_i} = \bigcup_{j\neq i}^{}{{P}_k^{c_j}}$. 
The target set of unlabeled vines with hash-tag $c_i$ are represented as $\mathbb{U}^{c_i}$. 
At $k^{th}$ iteration, the set of remaining unlabeled vines is $\mathbb{U}_k^{c_i}$, $\mathbb{U}_k^{c_i} \subset \mathbb{U}^{c_i}$. 

\paragraph{Initialization:}
The initial training of classifiers for class $c_i$ is performed using the auxiliary 
set (UCF50 examples) as the training set. At this point, all vines in the target set are unlabeled. 
Hence, for the $0^{th}$ iteration, 
\[\mathbb{T}_0 = \mathbb{A}, \,
\mathbb{P}_0^{c_i} = \mathbb{A}^{c_i}, \,
\mathbb{N}_0^{c_i} = \mathbb{A}^{\tilde{c_i}}, \,
\mathbb{U}_0^{c_i} = \mathbb{U}^{c_i}\]
For each class, we train the SVM classifiers $\mathrm{H_{FV}}$ and $\mathrm{H_{WV}}$ using samples from the initial training set $\mathbb{T}_0$. The SVM classifier for each class return a confidence score $\in [0,1]$ for each target sample. The two SVM scores are multiplied to yield a combined confidence 
score for every vine sample in the unlabeled target set $\mathbb{U}_0^{c_i}$. 
The multiplicative scoring function penalizes the overall score when any of the two scores is low and helps to ensure that only the samples with highest confidence are labeled. We pick the top-$K$ scoring vines as potential positive samples, where $K$ is emperically selected to be $10\%$ of the auxiliary positive set ($|\mathbb{P}_0^{c_i}|$) at every iteration. We update the negative set for a class $c_i$ by adding the newly labeled positives of other classes as labeled negatives for $c_i$. The auxiliary set size can be fixed or modified incrementally as explained later. The positives and negatives in iteration $k$ for the class $c_i$ are denoted as $\mathbb{L}_k^{c_i}$ and $\mathbb{L}_k^{\tilde{c_i}} = \bigcup_{j\neq i}{}{\mathbb{L}_k^{c_j}}$.
Note, we don't use tag based scoring in the initialization due to unavailability of tags for the initial auxiliary set. 

\paragraph{Iterative training and update:}
The training set for class $c_i$ at iteration $k > 0$ is formed by augmenting the newly labeled vines to the auxiliary set. 
\[\mathbb{T}_{k}^{c_i} = \mathbb{T}_{k-1}^{c_i} \cup \mathbb{L}_{k-1}^{c_i} \cup \mathbb{L}_{k-1}^{\tilde{c_i}}, \qquad \mathbb{P}_{k}^{c_i} = \mathbb{P}_{k-1}^{c_i} \cup \mathbb{L}_{k-1}^{c_i}, \qquad
\mathbb{N}_{k}^{c_i} = \mathbb{N}_{k-1}^{c_i} \cup \mathbb{L}_{k-1}^{\tilde{c_i}}\]
The parameters of the embedding function are re-estimated using 
the augmented training set. Our hypothesis is that by incrementally adding more vines to the training set,
in each iteration, we slowly modify the initial embedding that worked well for the auxiliary set to adapt for 
the target set (vines). As we are not altering the fisher vector space, we drop the motion feature classifier ($\mathrm{H_{FV}}$ ) after the initial iteration. The tag score for a target vine in iteration $k$ is the average number of co-occurring tags between given vine and positively labeled vines in the previous iterations. The tag-score $s_t$ is computed as follows,
\[s_t(x_t^v) = \frac{1}{|\mathbb{L}^{c_i}|}\frac{1}{\bar{n_t}}\,\sum_{x_p \in \mathbb{L}^{c_i}}\,\sum_{i=1}^{N_D}(x_t(i) \ast x_p(i)), \quad \text{where,} \quad \mathbb{L}^{c_i} = \mathbb{L}_{k-1}^{c_i} \cup \mathbb{L}_{k-2}^{c_i} \cup ... \mathbb{L}_{1}^{c_i}\]
Here, $N_D$ is the size of the tag dictionary and $\bar{n_t}$ is the average number of tags per vine in the target set (15 in our experiment). 
The combined score of a target training vine is computed by multiplying the semantic space SVM confidence scores and the tag-score. The
tag-score boosts the overall score of the test vines that have many co-occurring tags with the previously labeled
positive vines. The tags help in distinguishing samples of different classes retrieved as a result of the hash-tag. For
example, Apart from `diving', `sky-diving' and `diving in a pool' will have different accompanying tags which will match
accordingly to the currently classified/labeled vines. We stop the iterations when we have labeled approximately $50$\% of 
the auxiliary positives, i.e. ${P}_0^{c_i}$. 

\paragraph{Sampling choice for auxiliary set augmentation:}
In addition to augmenting the auxiliary set, we also perform an experiment where we gradually replace the auxiliary samples by target samples. This approach allows us to diminish the influence of auxiliary samples and provide more priority to target samples. We evaluate the performance of our method for both sampling choices, with and without replacement in \autoref{sec:experiment}. 

\section{Experiments and Results}
\label{sec:experiment}
In this section, we explain the experimental setup, establish the baseline performance, and report the results of our method. In the next section, we present analysis and interpretation of these results.

\label{sec:expsetup}

\subsection{Experimental Setup}
Here, we explain the code setup and parameters used for feature computation and classifier training. 
To extract the motion features, we compute the improved dense trajectory descriptors \cite{ImpDenseTraj,IDTjournal} for all videos using the code\footnote{\url{https://lear.inrialpes.fr/people/wang/improved_trajectories}} by the authors. For fisher vector computation, the experimental parameters 
are same as \cite{IDTjournal} and the GMM paramters are estimated over UCF samples of 50 action classes. The fisher vectors thus computed are of $101,376$ dimensions.
For computing the semantic features, we embed the fisher vectors into the $300$ dimensional word2vec space. 
For learning the embedding function, we use publicly available implementations of word2vec\footnote{\url{https://code.google.com/p/word2vec/}} 
and zero-shot learning \footnote{\url{https://github.com/mganjoo/zslearning}}. For initializing the embedding function we use $400$ hidden nodes and limit the maximum iterations to $1000$. For computing the dictionary of tag-words, we first perform stemming -- a commonly used 
trick in NLP applications to reduce the related forms of a word to its root form. 
From the remaining words we remove the singletons. The tag feature for a vine is a binary vector of the size 
of the tag dictionary such that each 0/1 element indicates whether that tag in the dictionary occurs with this vine or not. 
We use linear SVMs for fisher vector and semantic features classification in our iterative training. For SVMs we fix $C=1$ and 
the weight for the positive class to be $7$ times more than the negatives to compensate for fewer positive samples as compared to the negatives being added in each iteration. 


\subsection{Performance Evaluation}
\begin{figure}[h!]
 \centering
 \subfloat[ROC - Billiards]{\includegraphics[height=3.7cm]{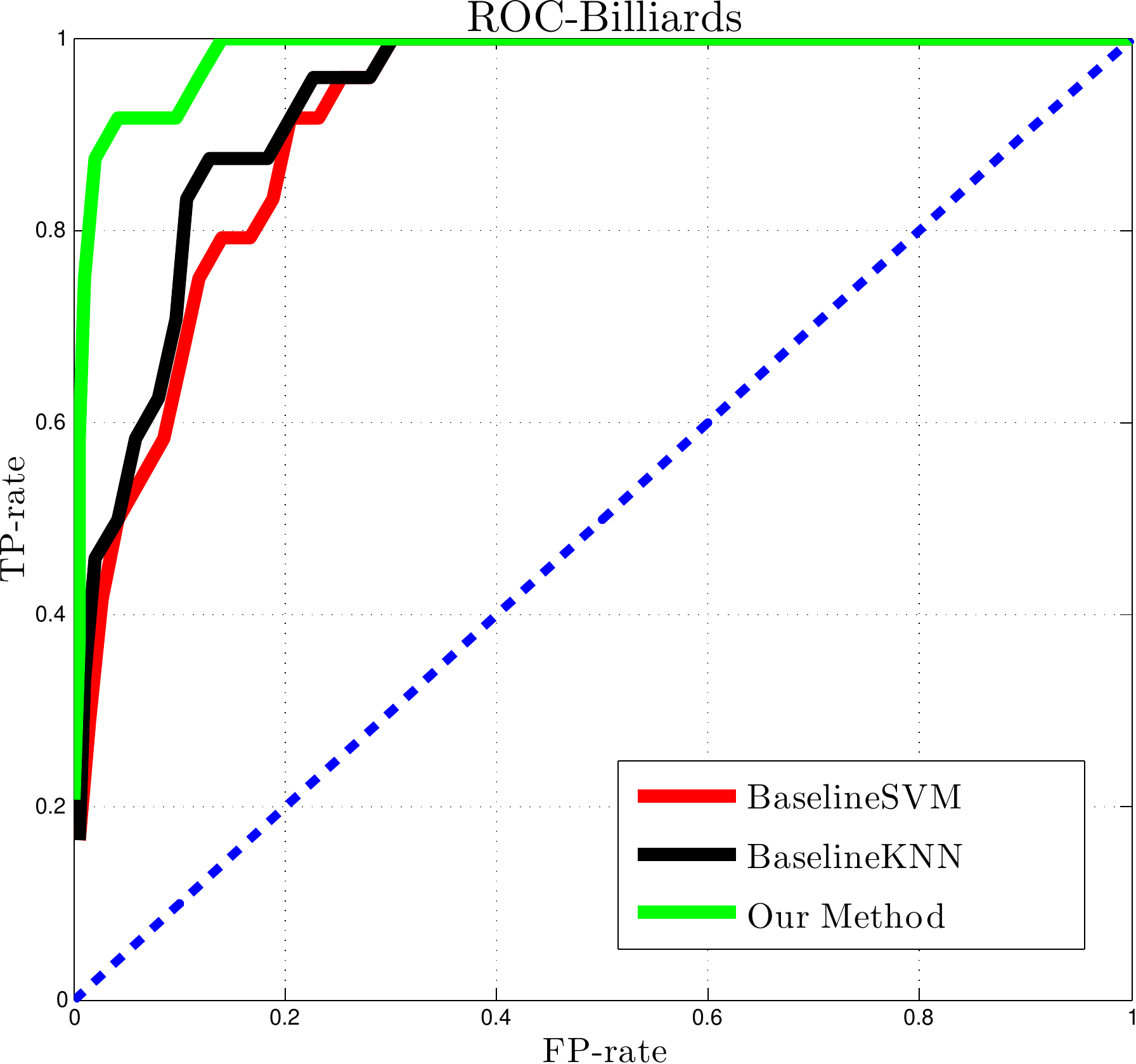}}\,
 \subfloat[ROC - Cycling]{\includegraphics[height=3.7cm]{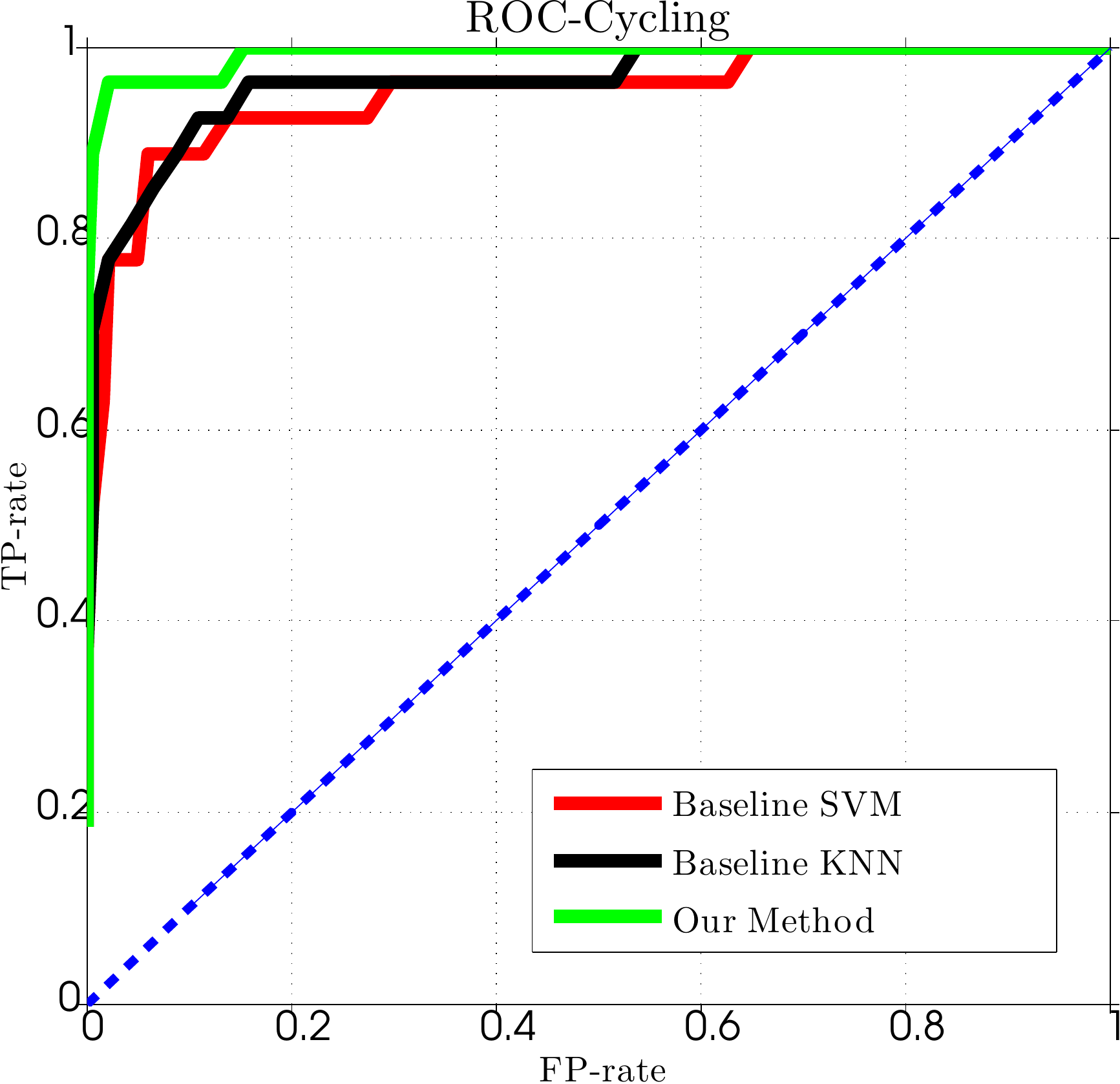}}\,
 \subfloat[ROC - Golf]{\includegraphics[height=3.7cm]{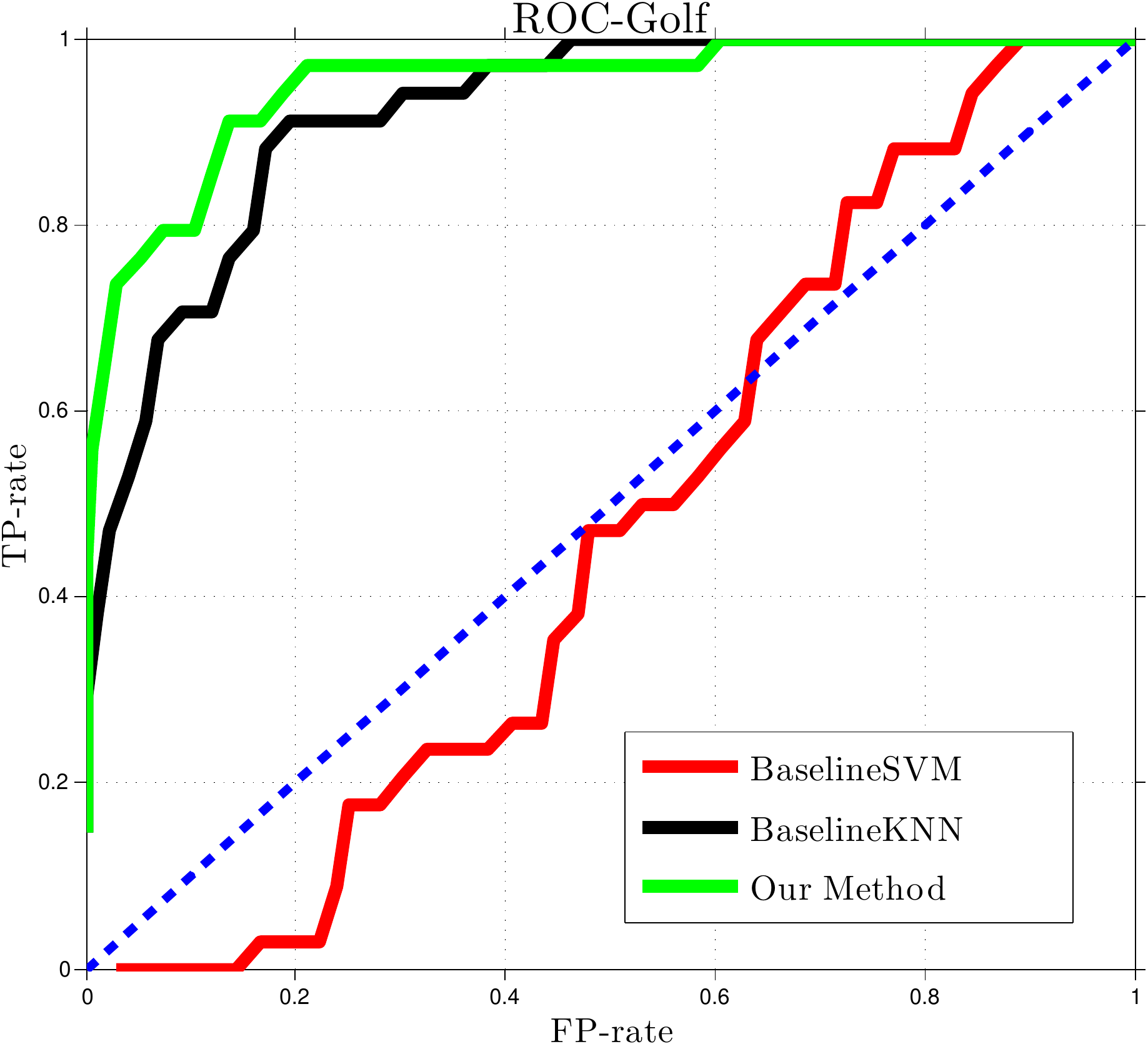}}
 \small{\caption{ROC comparison of our iterative training method (without replacement) with baseline FV+SVM and ES+NN.}\label{fig:roc}}
 \vspace{-2mm}
\end{figure}

We evaluate the performance of our approach by classifying a test set using the semantic embedding learned in the final iteration. 
We compare the performance of our method with the baseline classifiers trained on the auxiliary dataset
(UCF50) for semantic and motion features. We report precision, recall, and F-score for classification of 
the test dataset using all methods in \autoref{tab:res} and the ROC-curves for three classes 
are shown in \autoref{fig:roc}. The two baseline classifiers trained on the auxiliary dataset 
are represented as, (i) Motion-only (FV+SVM), and, (ii) Semantic-only (ES+NN). 

For Motion-only baseline, we train  $7$ one-vs-rest linear SVM classifiers and assign the labels based on the highest decision value of the corresponding classifiers. For Semantic-only baseline, we train an embedding function using the auxiliary set and use it to project the test samples to the semantic space. We classify the samples to the nearest class in the word2vec space using $L_2$ distance. The semantic embedding function learnt using our iterative approach is also evaluated in a similar fashion (ES+NN). We evaluate our final semantic embedding for both sampling choices , when (i) the auxiliary set videos are replaced by the newly labeled vines (with replacement), (ii) the auxiliary set is only augmented by the newly labeled vines (without replacement).

\begin{table*}[t]
\centering
\begin{adjustbox}{width=\textwidth}
\begin{tabular}{l c c c | c c c || c c c c c c}
\toprule 
& \multicolumn{6}{c}{Iterative Training(Ours)} & \multicolumn{6}{c}{Baseline Methods}\\
\cmidrule(lr){2-7}
\cmidrule(lr){8-13}
& \multicolumn{3}{c}{With-replacement} & \multicolumn{3}{c}{Without-replacement} & \multicolumn{3}{c}{FV\textunderscore svm} &\multicolumn{3}{c}{ES\textunderscore knn} \\
        \cmidrule(lr){2-4}
        \cmidrule(lr){5-7}
        \cmidrule(lr){8-10}
        \cmidrule(lr){11-13}
         Class & prec.\ & rec.\  & F-score\ & prec.\  & rec.\ & F-score\  & prec.\ & rec.\  & F-score\ & prec.\ & rec.\
         & F-score\ \\
         \midrule
         Billiards   &.750&.875&.807&73.3&.916&\textbf{.814}&1.00&.166&.285&1.00&.250&.400\\
         Cycling     &.956&.814&\textbf{.880}&.884&.851&.867&.585&.888&.705&.621&.851&.718\\
         Diving      &.750&.538&.626&.814&.564&\textbf{.666}&.645&.512&.571&.620&.461&.529\\
         Golf-Swing  &.909&.588&.714&1.00&.588&\textbf{.740}&.641&.735&.684&.638&.676&.657\\
         Horseriding &.634&.896&.742&70.2&.896&.787&.857&.827&\textbf{.842}&.857&.827&\textbf{.842}\\
	 Kayaking    &.388&.875&\textbf{.538}&.361&.812&.500&.407&.687&.511&.333&.750&.461\\
	 Pushups     &.967&.750&.845&.969&.800&\textbf{.876}&.846&.825&.835&.891&.825&.857\\
         \bottomrule
\end{tabular}
\end{adjustbox}
\caption{Comparison table for our methods with the baselines.}
\vspace{-4mm}
\label{tab:res}
\end{table*}

\section{Analysis}

\begin{figure}[t]
\centering
\fbox{\subfloat[]{\includegraphics[height=3cm, width=0.45\linewidth]{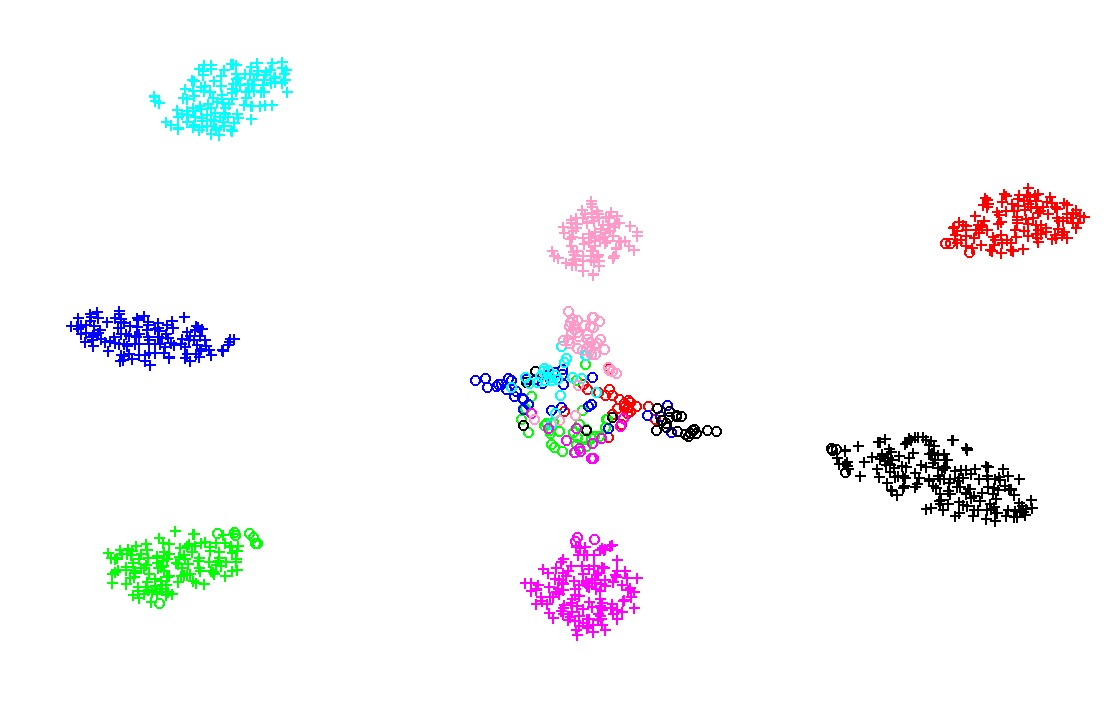}}}
\fbox{\subfloat[]{\includegraphics[height=3cm, width=0.45\linewidth]{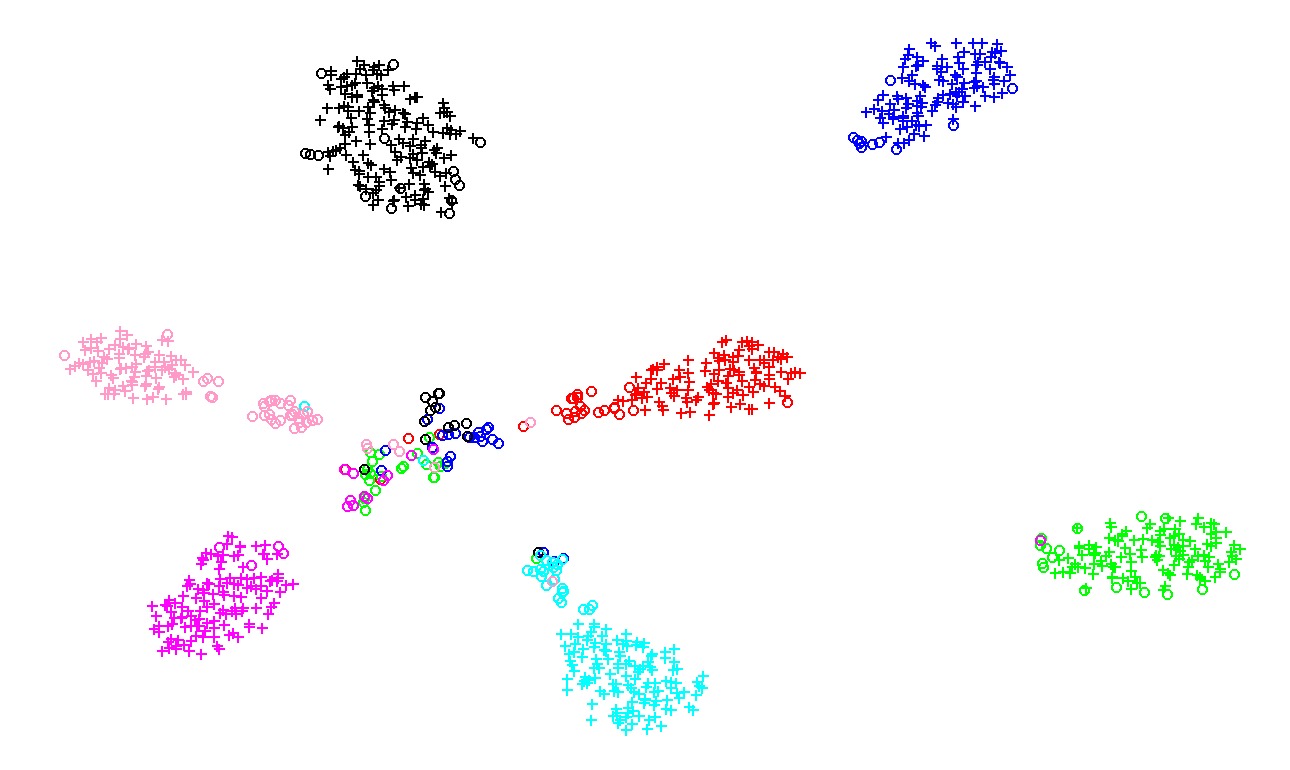}}}
\caption{t-SNE Visualization of semantic embedding of UCF and Vines before and after iterative training. This visualization is best viewed in color.}
 \label{fig:vis_test_distributions}
 \vspace{-3mm}
\end{figure}

\paragraph{Iterative relearning of embedding function}
In \autoref{fig:vis_test_distributions}, we show the effect of relearning the embedding function by t-SNE visualization of the test samples. In $0^{th}$ iteration, we see UCF samples forming separate clusters but vines forming an inseparable distribution. This shows that the embedding function learnt solely using auxiliary set is insufficient to classify the test set. However, after iterations we see that the embedding function projects most of the test samples around their class labels and close to the corresponding UCF samples. This observation supports our hypothesis of incrementally modifying the embedding function using augmented training set. If this hypothesis was faulty then the embedding function would degrade performance of the test classification after iterations (see \autoref{fig:vis_test_distributions}, \autoref{tab:res}). In supplementary material we show additional details and visualizations per iteration.    

\paragraph{Comparision of baseline classifiers}
\autoref{tab:res} shows that the baseline methods, FV+SVM and ES+NN, have comparable performances. However, ES+NN operates in a significantly lower dimensional space ($101376$ vs.\ $300$) and has much lower time complexity serving as a better alternative.

\paragraph{Without-replacement vs.\ baseline classifiers}
This sampling method outperforms both baseline methods for all classes except `Horseriding' \& `Kayaking'. For Horseriding we obtain a higher recall, however due to fall in precision the overall performance decreases. Though our method performs better than ES+NN but falls short on precision against FV+SVM. Kayaking contains the least ratio of true positives ($< 25\%$) in the target set (\autoref{tab:base_datasets}). Incremental addition of target samples still helps in improving precision and recall as compared to ES+NN baseline for `kayaking'.

\paragraph{With-replacement vs.\ baseline classifiers}
This sampling method performs significantly better for all classes except `Horseriding'. For `Kayaking' our method performs marginally better and as mentioned earlier the improvement is limited due to the lack of positives in the target dataset. For `Billiards', we achieve an acceptable precision with significantly higher recall against both the baselines which show a highly skewed performance.

\paragraph{With-replacement vs.\ without-replacement}
We experimented with these two sampling choices to see whether diminishing the influence of auxiliary domain will hamper or aid the process of learning the embedding function. It is evident from \autoref{tab:res}, for 5 of the 7 classes without-replacement perfoms better and for the other two classes the difference is marginal. This result suggests that augmenting the auxiliary domain by target domain without replacement has a positive influence on iterative learning.

\section{Conclusion \& Future Work}
\label{sec:conclusion}
In this paper we explored the problem of improving the performance of action classification for an unseen unlabeled wild domain by utilizing labeled examples from a simpler source domain. We presented a simple iterative technique that improves semantic embedding of video features into a structered reference space to bring together the disparate auxiliary and target domains. We showed the effectiveness of this iterative technique on the task of action classification for $7$ classes in vines. We also experimented with two sampling choices for augmenting the auxiliary domain and presented detailed analysis of results. In future, we would like to explore the application of this method for annotation and tag enrichment applications. 

%
%

{
	\bibliographystyle{plainnat}
	\small{\bibliography{egbib}}
}

\end{document}